\documentclass[lettersize,journal]{IEEEtran}
\usepackage{amsmath,amsfonts}
\usepackage{algorithmic}
\usepackage{algorithm}
\usepackage{array}
\usepackage[caption=false,font=normalsize,labelfont=sf,textfont=sf]{subfig}
\usepackage{textcomp}
\usepackage{stfloats}
\usepackage{url}
\usepackage{verbatim}
\usepackage{graphicx}
\usepackage{booktabs}
\usepackage{cite}
\usepackage{makecell}
\usepackage{multirow,graphicx}
\usepackage[dvipsnames]{xcolor}

\newtheorem{remark}{Remark}

\usepackage{comment}

\usepackage{url}

\newcommand{\T}{\mathcal{T}}
\newcommand{\N}{\mathcal{N}}

\newcommand{\I}{\mathcal{I}}
\newcommand{\generators}{\mathcal{I}_n}
\newcommand{\segments}{\mathcal{S}_i}

\newcommand{\demand}{\mathrm{D}_n}
\newcommand{\linepacacity}{\mathrm{F}_{nm}}
\newcommand{\segbound}{\mathrm{Q}_s}

\newcommand{\bid}{\mathrm{c}_{i,s}}
\newcommand{\bidvector}{\boldsymbol{c}_i[t]}

\newcommand{\qseg}{q_{i,s}}
\newcommand{\busangle}{\phi_{n}}
\newcommand{\flow}{f_{nm}}
\newcommand{\dispatch}{q_i^*[t]}

\begin{document}

\title{AI agents in Algorithmic Electricity Markets: \\On the Emergence of Tacit Collusion}

\author{Jakub~Seredyński,~Georgios Tsaousoglou

\thanks{\noindent \hspace{-8pt}The authors are with the Department of Applied Mathematics and Computer Science, Technical University of Denmark.\\ e-mail correspondance: geots@dtu.dk.\\
This work was supported by the Villum Young Investigator award (Project no. 80181).\\
}
}

\maketitle

\begin{abstract}
As electricity market participants increasingly adopt learning-based agents for their bidding strategies, electricity markets are becoming \textit{algorithmic}.
Evidence from algorithmic markets in other domains shows that tacit collusion can arise purely through independent learning.
Moreover, electricity markets are typically oligopolistic and feature repeated interaction among a small number of participants, making them structurally susceptible to non-competitive behavior.
In the face of these observations, this paper investigates the hypothesis that tacit collusion may emerge in electricity markets where participants' actions are controlled by autonomous learning-based algorithms. We model strategic bidding as a repeated game with imperfect public monitoring, and model the participants' emergent behavior using multi-agent reinforcement learning.
We propose a multi-dimensional set of criteria (going beyond profit comparisons against Nash equilibria) to assess whether the resulting behavior constitutes tacit collusion. Our experimental results showcase that such a danger is realistic for electricity markets: there are cases where agents do learn to sustain supra-competitive outcomes that are supportive of tacit collusion indicators, even though the agents were never instructed to collude.
\end{abstract}

\noindent


\section{Introduction}
\subsection{Motivation and Research Question} \label{motivation}

\IEEEPARstart{C}{ollusion} in electricity markets refers to an agreement among competing firms to coordinate their actions towards avoiding competition and increasing their profits. It is severely detrimental, not only to consumers who face supra-competitive prices, but also to the market's economic efficiency in general. 
While collusion is associated with cartel-like practices and is thereby strictly prohibited by law, the more challenging case to deal with is when collusion is \textit{tacit}.

\noindent \textit{Definition:} \noindent\textbf{Tacit Collusion} refers to a group of oligopolists’ ability to coordinate, even in the absence of explicit agreement, in order to raise prices or, more generally, increase profits at the detriment of consumers \cite{ivaldi2007economics}. \hfill $\square$

Importantly, the absence of a need for an explicit agreement makes it possible for Tacit Collusion to emerge ``autonomously''.
This is directly relevant to the new reality of electricity markets where participants are increasingly outsourcing their market participation strategies to autonomous learning-based algorithms, generally referred to as Artificial Intelligence (AI) agents. Experience from other domains has already established that AI agents, particularly Reinforcement Learning (RL), often exhibit the ability to maximize their prescribed reward (read: profit) in ways unforeseen by their designers.
The study in \cite{calvano2020artificial} demonstrates that independently designed reinforcement-learning pricing algorithms can indeed learn to sustain supracompetitive prices, in general contexts. In the authors' own words: ``{We find that the algorithms consistently learn to charge supracompetitive prices, without communicating with one another [...] The algorithms learn these strategies purely by trial and error. They are not designed or instructed to collude, they do not communicate with one another, and they have no prior knowledge of the environment}''.

Taken to electricity markets, such agents could give rise to Tacit Collusion phenomena, even without the firms' intention or awareness.
Moreover, the fact that electricity markets historically tend to be oligopolistic while also featuring repeated interaction among participants, makes them particularly vulnerable to Tacit Collusion emergence.
Motivated by the possibility and implications of this phenomenon in electricity markets, we put forward the following hypothesis:

\noindent\textbf{Hypothesis}: \textit{In an electricity market where the participants' actions are controlled by autonomous learning-based algorithms, Tacit Collusion may emerge.}

\subsection{Related Work}

An electricity market comprises participants, each of which is interested in optimizing its own profit. Thereby, studying possibilities over electricity market outcomes involves modeling the bidding strategy optimization problem of individual participants, in an agent-based modeling fashion \cite{liang2020agent}. 
In the earlier literature, a participant's bidding strategy was predominantly modeled 
through bi-level programming (e.g. \cite{kardakos2014optimal}), giving rise to mixed-complementarity problems modeling the multi-participant interactive decision-making \cite{hu2007using, gabriel2013complementarity}.
The recent literature, however, has substantially pivoted toward learning-based approaches, for at least three reasons:
\begin{itemize}
    \item Learning algorithms are able to handle realistic aspects of electricity markets such as complexity, partial observability, and stochasticity, departing from simplifying assumptions arising from computational constraints of the model \cite{tsaousoglou2022market}.
    \item The model-free nature of learning algorithms makes them a generic modeling tool which, taken to a sufficient level of richness, can in principle reproduce a wide range of different intelligent strategies \cite{tang2024forecasting, baltaoglu2018algorithmic}.
    \item Real market participants increasingly experiment with learning-based bidding strategies \cite{eschenbaum2026shared}. This motivates the RL framework, in particular, to be leveraged as a suitable modeling framework that captures the exploration aspect of modern algorithmic trading.
\end{itemize}

RL has attracted special interest as a generic model for a participant's bidding strategy optimization problem \cite{ye2019deep}, \cite{zhu2023reinforcement}, \cite{hu2025strategic}. 
From a system perspective, when it comes to modeling the interactive learning among several strategic participants and the emergent outcomes thereof, the concept naturally evolves into multi-agent RL (MARL) \cite{zhang2025taming}.
To that end, \cite{harder2023fit} argues for MARL-based fit-for-purpose simulators that retain the market and physical features relevant to the research question at hand, while \cite{du2021approximating} formulates day-ahead bidding with multiple strategic generators as algorithmic traders and applies a MARL algorithm to approximate Nash-equilibrium bidding.
Recent work on generator modeling further shows that simplified bidding and marginal-cost representations can affect conclusions in market-mechanism simulations, especially when renewable integration changes the operating regimes of thermal \mbox{generators~\cite{pan2025decision}.}


The potential emergence of Tacit Collusion has been validated by research in other domains, namely algorithmic pricing, where RL agents have been shown to sustain supra-competitive prices without communication or explicit collusive instructions~\cite{calvano2020artificial}. This result has been extended to sequential pricing environments, where Q-learning algorithms may converge either to collusive equilibria or to supra-competitive price cycles~\cite{klein2021autonomous}. Recent experimental evidence further compares algorithmic and human collusion in the same market environments, showing that self-learning pricing algorithms can generate more collusive outcomes than human decision-makers in some oligopoly settings~\cite{werner2021algorithmic}.



Taken to the electricity-market setting, it is particularly timely to understand the implications of algorithmic trading and, in particular, whether Tacit Collusion can emerge endogenously from the interaction of autonomous learning agents. To that end, a MARL agent-based model is employed in \cite{liang2020agent}, where the authors investigate how the agents' discount factors affect the resulting market outcomes. More recently, \cite{aliabadi2022emerging} explicitly investigates the emergence of Tacit Collusion and define a collusive state as one in which the profit of each participant exceeds that attained under all Nash equilibria. While this provides a useful benchmark for identifying supra-competitive outcomes, establishing such a condition requires the computation of the Nash equilibria of the underlying market game. This requirement substantially constrains the complexity of the market model and makes the approach difficult to extend to richer electricity-market settings.

\subsection{Research Gap and Contributions}

The emergence of algorithmic trading, and the risks it poses for electricity markets, points to a major research need around the emergence of Tacit Collusion. To our knowledge, this phenomenon has been addressed by only a handful of studies (namely \cite{aliabadi2022emerging} and \cite{liang2020agent}).
These studies provide important evidence that intelligent agents may learn supra-competitive bidding strategies. Their identification of Tacit Collusion, however, is closely tied to the structure of the underlying market game. 
In particular, \cite{aliabadi2022emerging} identifies collusion by comparing the agents' profits with those attained at all Nash equilibria.
Such an approach is informative in a stylized market model whose equilibria can be fully characterized, but becomes difficult to apply once the market model incorporates the structural features of actual electricity markets.

More fundamentally, supra-competitive profits alone do not establish Tacit Collusion: elevated profits can arise for reasons unrelated to collusive coordination. Conversely, because realistic market models rarely converge to a stage-game Nash equilibrium, the absence of such convergence is, on its own, equally uninformative about whether collusion has occurred. In fact, in our numerical results in this paper, we present examples on both ends: cases where supra-competitive outcomes emerge which are, or are not necessarily, supportive of a case for Tacit Collusion.
Tacit Collusion is thus a subtle phenomenon that resists identification through any single criterion and calls instead for a multi-dimensional examination. This motivates the contributions of this paper, positioned towards investigating the Hypothesis defined in Section \ref{motivation}:

\begin{enumerate}
    \item We adopt Tacit Collusion criteria (beyond merely supra-competitive outcomes) from other domains, with richer experience in this context, and adapt them to the electricity market setting. We also propose a novel criterion which quantifies the central characteristic of collusion as defined in its \textit{Definition} (cf Section \ref{motivation}), in the context of electricity markets.
    \item We model emergent bidding profiles using a MARL framework and examine their collusive properties across multiple such criteria.
    \item We experimentally investigate which market characteristics (e.g. grid constraints' tightness, demand level, marginal costs' variance) enhance the possibility of emergent collusion.
\end{enumerate}


\noindent The remainder of the paper is organized as follows.
Section \ref{sec:model} presents the system model. Section \ref{sec:tacit_collusion} defines the different Tacit Collusion criteria of our study. Section \ref{sec:experimental_setting} presents the experimental setting, including the details of the MARL algorithm. Section \ref{sec:results} presents our experimental results, while Section \ref{sec:conclusion} concludes the paper.

\section{System Model}\label{sec:model}

\subsection{Electricity Market}
We consider a single-timeslot electricity market, operating over a set $\N$ of buses, which is cleared by solving the standard DC optimal power flow (OPF) problem.
Specifically, a bus $n \in \N$ features an inelastic demand $\demand$ and a set $\generators$ of energy-providing resources. The resources can generally be of various types and technologies (including conventional power plants, renewable energy sources, storage, flexible demand, etc) but we will simply call them ``generators'' to simplify the exposition. 
We denote the superset of generators (across all nodes) by $\I = \bigcup_{n\in \N} \generators$.

The active power flow from bus $n$ to bus $m$ is denoted as $\flow$ and is subject to line capacity bounds per
\begin{equation} \label{c-flows}
    -\linepacacity \leq f_{nm} \leq \linepacacity, \quad \forall (n,m) \in \N^2,
\end{equation}
where, for pairs of buses that are not connected by a line, we simply set $\linepacacity = 0$.
Denoting the output of a generator $i \in \generators$ as $q_i$, the power balance constraint at a node reads:
\begin{equation} \label{c-powerbalance}
    \sum_{i \in \generators}q_i - \sum_{m \in \N} f_{nm}  = \demand, \quad \forall n \in \N.
\end{equation}
The phase angle at a bus $n$ is denoted as $\busangle$ and is subject to safety limits per
\begin{equation} \label{c-angles}
    \underline{\phi}_{n} \leq \busangle \leq \overline{\phi}_{n}, \quad \forall n \in \N.
\end{equation}
The phase angles and the power flows are linked by the standard DC power flow model:
\begin{equation} \label{c-dc}
    f_{nm} = \mathrm{B}_{nm}(\phi_{n} - \phi_{m}), \quad \forall (n,m) \in \N^2,
\end{equation}
where $\mathrm{B}_{nm}$ is the susceptance of transmission line $nm$.

The output of a generator $i \in \I$ is partitioned into a set $\segments$ of segments, as in:
\begin{equation} \label{c-segsum}
    q_i = \sum_{s \in \segments} \qseg, \quad \forall i\in \I,
\end{equation}
where the output $\qseg$ of a segment is bounded by
\begin{equation} \label{c-seg}
    0 \leq \qseg \leq \segbound, \quad \forall s \in \segments, i\in \I.
\end{equation}

Finally, each generator declares a per-unit energy cost $\bid$ for each of its segments $s \in \segments$, such that the operator receives a piecewise linear cost function for each generator.
The market is cleared by solving the standard DC-OPF problem:
\begin{align} 
    \underset{q_i, \qseg, \flow, \busangle}{\text{min}} ~ \Big \{ 
   \sum_{i \in \I} \sum_{s \in \segments} &
   \bid \qseg \Big\} \label{market-clearing}\\
    \text{subject to}& \notag\\
    \eqref{c-flows} - \eqref{c-dc}&: \text{Network constraints} \notag\\
    \eqref{c-segsum}-\eqref{c-seg}&: \text{Generators' constraints}. \notag
\end{align}
The optimal dual variable $\lambda_n$ of the power balance constraint \eqref{c-powerbalance} instantiates the Locational Marginal Price (LMP) of \mbox{node $n$.}
The market-clearing problem \eqref{market-clearing} is implemented sequentially for a set of instances (timeslots) $\T$ where in each instance $t \in \T$ the generators can submit different bids $\bid[t]$ resulting in timeslot-specific solutions $\big((\dispatch)_{i \in \I}, (\lambda_n[t])_{n \in \N} \big)$. 

\subsection{Agent-based Market Participation Strategies}
A generator's market participation strategy is instantiated by an agent.
We use the terms ``agent'' and ``generator'' interchangeably and we index either by $i$.
After each market-clearing instance $t$, the stage profit $\pi_i[t]$ of $i$ is
\begin{equation} \label{profit}
    \pi_i[t] = \lambda_{n(i)}[t] \dispatch - C_i(\dispatch).
\end{equation}
The first term is the generator's revenue, with $\lambda_{n(i)}[t]$ being the LMP at bus $n(i)$ where $i$ is located.
The second term is the generator's real cost for generating its dispatched quantity $\dispatch$, which is given by the generator's cost \mbox{function $C_i: \mathbb{R} \rightarrow \mathbb{R}$.}

The values for $\dispatch$ and $\lambda_{n(i)}[t]$ are given from problem \eqref{market-clearing}, and are therefore affected by the \textit{actions}, i.e. bids, $(\bid[t])_{s \in \segments}$ of the focal agent $i$, as well as by those of the other agents.
This directly leads to the definition of the stage game $\mathcal{G}$:
\begin{itemize}
    \item \textbf{Players:} $i \in \I$
    \item \textbf{Actions:} $\bidvector = (\bid[t])_{s\in \segments}$
    \item \textbf{Payoffs:} $\pi_i[t]$ as in Eq. \eqref{profit}
\end{itemize}

We consider agents that learn to improve their strategies through experience.
After each instance, an agent gets to observe its own dispatch and profit, as well as the resulting LMP at every node of the network:
\begin{equation} \label{observation}
    o_i[t] = \big(\dispatch, \pi_i[t], (\lambda_n[t])_{n \in \N}\big).
\end{equation}
This gives rise to a repeated game with imperfect monitoring, since an agent does not observe the actions or profits of other agents but only a public signal (i.e. the LMPs) from which the others' actions and profits cannot be inferred.
Furthermore, we do not assume that agents know the structure of the network or even the market mechanism and consider agents that learn purely through experience. To that end, each agent accumulates a private history of its actions and observations (up to $\zeta$ episodes behind):
\begin{equation} \label{history}
    h_i[t] = \big(\boldsymbol{c}_i[t-1], o_i[t-1], \boldsymbol{c}_i[t-2], o_i[t-2], ..., \boldsymbol{c}_i[t-\zeta], o_i[t-\zeta]   \big) 
\end{equation}
and selects its bid at each stage $t$ according to a \textit{strategy} $\sigma_i$ that maps the current history $h_i[t]$ to its decision $\bidvector$. We denote the joint strategy as $\boldsymbol{\sigma} = (\sigma_i)_{i \in \I}$.
Based on the above, we can define the objective of an agent as a search for a strategy that maximizes its cumulative profits over time:
\begin{align} 
    &\underset{\sigma_i}{\text{max}} ~ \Big \{ 
   \mathbb{E}_{\boldsymbol{\sigma}} \sum_{t = 0}^{\infty} \pi_i[t] \Big\} \label{agent-objective}\\
    &\text{subject to} \notag\\
    &\eqref{profit}: \text{Profits depend on actions},\notag\\
    &\bidvector = \sigma_i(h_i[t]): \text{Actions are given by strategy and history}\notag\\
    &\eqref{history}: \text{The history stacks part actions and observations}\notag\\
    &\eqref{observation}: \text{Observations are:  dispatch, profit, and LMPs}. \notag
\end{align}

Problem \eqref{agent-objective} is intractable in general: each agent optimizes its own strategy while the strategies of all other agents are unknown and simultaneously evolving, making the environment non-stationary from any single agent's perspective.
In line with the learning-through-experience consideration, we adopt \textit{Multi-Agent Reinforcement Learning} (MARL) as a model of how AI agents learn and play in practice.
%
%
At each instance $t$, every agent $i$ selects a bid $\bidvector = \sigma_i(h_i[t])$, the market clears, and each agent receives the observation $o_i[t]$ and uses it to update its strategy $\sigma_i$.
This process is iterated over successive instances, and the agents' strategies co-evolve. The details of the MARL algorithm will be presented in Section \ref{marl-algorithm-section}.

The output of this process is a joint strategy profile $\boldsymbol{\sigma}^* = (\sigma_i^*)_{i \in \I}$, representing the emergent bidding behaviour that arises when self-interested agents learn to bid. 
The paper's \textit{Hypothesis} thereby refers to analyzing the emergent joint strategies $\boldsymbol{\sigma^*}$ in terms of indications for Tacit Collusion. The next section formalizes the Tacit Collusion indices to be used.

\section{Tacit Collusion}
\label{sec:tacit_collusion}

In this section, we define three criteria that constitute indicators of the emergent joint strategy $\boldsymbol{\sigma^*}$ being tacitly collusive: Punishment of Deviation, Unsustainability under Shortsight, and Short-term Profitability of Deviations. 
In the three subsection below, we describe how each criterion is instantiated in our electricity market context.

\subsection{Punishment of Deviation Criterion}
This criterion checks whether an agent that deviates from the joint strategy $\boldsymbol{\sigma^*}$ and reverts to an explicitly competitive strategy will face a punitive response by other agents.

A deviator reverting to a competitive strategy means that it starts to optimize its short-term profits aggressively.
In our context, we can model such a situation by choosing one agent $j$ as the deviator and, after the MARL strategies have converged to $\boldsymbol{\sigma^*}$, switching the deviator's strategy to the one that optimizes its stage profit under the assumption that other agents $i \neq j$ continue to play their previous strategies (specifically, their most recently observed ones).
To implement this, at each after-convergence stage $t$, we force $j$'s strategy to the solution of the following bi-level optimization problem:
\begin{align} 
    &\underset{\boldsymbol{c}_j[t]}{\text{max}} ~ \Big \{ \lambda_{n(j)}[t] q^*_j[t] - C_j(q^*_j[t]) \Big\} \label{mpec}\\
    &\text{subject to} \notag\\
    &q^*_j[t], \lambda_{n(j)} \in \eqref{market-clearing}
    \notag
\end{align}
where the constraint specifies that the dispatch and LMP result through the market clearing problem \eqref{market-clearing}.
Problem \eqref{mpec} can be solved by casting the lower-level problem \eqref{market-clearing} as a set of KKT conditions and introducing auxiliary binary variables to handle the resulting complementarity constraints. This results in a single-level mixed-integer linear program which can be tackled by commercial solvers. The full derivation is ommited here as it is fairly standard in the literature.

While the deviator is using its aggressive profit-maximizing strategy \eqref{mpec}, the other agents continue to play by the MARL regime. A punitive response is considered to be one that has other agents lowering
their mark-ups in a way that reduces the deviator’s profit.
The test is considered positive if the post-deviation dynamics display the qualitative pattern of punish-and-forgive strategies:
\begin{itemize}
    \item Immediately after the deviation, competitors temporarily switch to more competitive (i.e. lower) bids, reducing prices \textit{and} the deviator’s market share;
    \item After a punishment phase, competitors gradually return to their pre-deviation strategies, while prices recover.
\end{itemize}

\subsection{Unsustainability under Shortsight Criterion}
Two elements that generally facilitate Tacit Collusion are:
\begin{itemize}
    \item agents playing the long game, by adopting a high discount factor which increases the importance of future payoffs over immediate ones;
    \item agents having memory of past actions and rewards, which allows them to condition their behaviour on history;
\end{itemize}
The rationale of this criterion is that reducing or eliminating these elements should make it substantially more difficult for agents to learn coordinated strategies. In practice, this test is implemented by executing the MARL algorithm under a very low discount factor while also severely truncating the memory of past
actions and rewards. Compared to the baseline case, this run would presumably result in more competitive strategies, i.e. lower prices and profits.


\subsection{Short-term Profitability of Deviations Criterion}
Given that collusive outcomes are not Nash equilibria of the stage game, one or more agents do have a profitable unilateral strategy deviation.
Yet the collusion is sustained when agents refrain from switching to opportunistic strategies, despite this being profitable in the short-run, presumably because they understand that those would lead to down-the-road less profitable outcomes.
This criterion checks the emergent joint strategy $\boldsymbol{\sigma}^*$ precisely for these two conditions: agents having profitable unrealized deviations from it, and whether those, if realized, would trigger joint adaptation dynamics that indeed lead to less profitable trajectories. 

Similarly to the first criterion, we instantiate profitable deviations by solving the bi-level problem \eqref{mpec}. This time, however, we solve problem \eqref{mpec} iteratively and for every agent in turn, thereby simulating a relapse into competitive (non-collusive) strategy adaptation.
The relevant algorithm is called ``Iterative Best Response'' (IBR) or ``diagonalization'' and it is widely used in the electricity markets' literature (see e.g. \cite{avila2025inefficient} and references therein) to approximate Nash Equilibria under the so-called Equilibrium-Problem with Equilibrium-Constraints (EPEC) model.
The criterion considers that $\boldsymbol{\sigma}^*$ passes the collusion test if, upon switching from MARL to the IBR regime, the profits of the first deviating agents:
\begin{itemize}
    \item increase at first, and
    \item they eventually drop below those under $\boldsymbol{\sigma}^*$, as the IBR process continues towards a Nash Equilbirium.
\end{itemize}
The rationale is that, if both of these conditions hold true, it means that each agent has a profitable unilateral deviation which it doesn't realize, possibly because it foresees that it is not profitable under the other agents' corresponding competitive response, thereby staying true to a coordinated (collusive) strategy.

\section{Experimental Setting}
\label{sec:experimental_setting}

\subsection{Electricity Market Setup}


We conduct the experiments on a stylised seven-bus electricity market adapted from the case study of \cite{aliabadi2022emerging}. The original system is an extended version of the well-known Pennsylvania--New-Jersey--Maryland (PJM) five-node test system and was designed as a compact networked market in which strong collusive equilibria can arise. We use it here as a controlled benchmark that preserves the elements needed for our hypothesis: an oligopolistic supply side, inelastic nodal demand, transmission constraints, and locational marginal prices.

The market contains four strategic generators located at buses $N_1$, $N_2$, $N_5$, and $N_6$. Nodal demand is perfectly inelastic and the market is cleared, at each instance, by the DC-OPF problem \eqref{market-clearing}. The resulting dispatch and nodal prices determine each generator's stage profit according to \eqref{profit}, where revenues are computed at the LMP of the generator's bus and costs are computed using the generator's true marginal cost. The main generation and cost parameters used in the experiments are reported in Table~\ref{tab:market_setup_generators}.

\begin{table}[t]
\caption{Strategic generators and marginal-cost cases.}
\label{tab:market_setup_generators}
\centering
\scriptsize
\newcolumntype{C}{>{\centering\arraybackslash}p{0.13\columnwidth}}
\setlength{\tabcolsep}{0pt}
\begin{tabular*}{\columnwidth}{@{\extracolsep{\fill}}lccCCC@{}}
\toprule
Generator & Bus & $P_i^{\max}$ [MW] & \multicolumn{3}{c}{Marginal cost [EUR/MWh]} \\
\cmidrule(lr){4-6}
 & & & Low & Medium & High \\
\midrule
$G_1$ & $N_1$ & 42 & 20 & 20 & 20 \\
$G_2$ & $N_2$ & 35 & 20 & 20 & 20 \\
$G_5$ & $N_5$ & 40 & 20 & 30 & 40 \\
$G_6$ & $N_6$ & 40 & 20 & 10 & 1 \\
\bottomrule
\end{tabular*}
\end{table}

All exogenous fundamentals are kept fixed within a scenario. In particular, nodal demand, marginal costs, and network parameters do not vary over time. The three demand cases, indexed by buses $(N_1,\ldots,N_7)$, are $D^{L}=(17,10,8,6,13,0,12)$, $D^{M}=(22,14,10,8,18,0,13)$, and $D^{H}=(26,15,12,9,20,0,14)$. Thus, the temporal dynamics observed in the simulations are induced by the repeated interaction of bidding strategies and by the feedback of the market-clearing mechanism, rather than by exogenous factors.

To isolate the role of transmission constraints, we consider two network representations. In the \texttt{Grid} case, the network constraints \eqref{c-flows}--\eqref{c-dc} are enforced using the topology and line limits of the test system. In the \texttt{NoGrid} case, the same demand and generator data are retained, but line capacities are set sufficiently high so that the bounds in \eqref{c-flows} never become binding, while the corresponding admittance values are chosen so that the DC flow relation \eqref{c-dc} and phase-angle bounds \eqref{c-angles} do not restrict the optimal dispatch. Operationally, this produces a copper-plate counterfactual. Mathematically, for all solved instances, the \texttt{NoGrid} case is equivalent to relaxing the network-induced restrictions in \eqref{c-flows}, \eqref{c-angles}, and \eqref{c-dc}, so that problem \eqref{market-clearing} clears the market as a single unconstrained zone.

The experimental design is a full factorial combination of three axes: network representation ${\mathrm{\texttt{Grid}} / \mathrm{\texttt{NoGrid}}}$, demand level ${\mathrm{\texttt{Low}}/ \mathrm{\texttt{Medium}} / \mathrm{\texttt{High}}}$, and marginal-cost heterogeneity ${\mathrm{\texttt{Low}} / \mathrm{\texttt{Medium}} / \mathrm{\texttt{High}}}$. This gives $2\times 3\times 3=18$ market environments. The selected scenario dimensions correspond to market characteristics that are widely discussed in the Tacit Collusion literature as factors influencing the emergence and sustainability of collusive outcomes, namely network topology, demand conditions, and cost heterogeneity~\cite{ivaldi2007economics,thal2011optimal}. All scenarios are trained and screened, while the behavioural collusion tests are applied only to the most informative cases selected according to the procedure described next.


\subsection{Scenario Selection Methodology}

Instead of running our tests in randomly selected market instances, we identify instances that are most suspicious for potential Tacit Collusion emergence.
In this subsection, we describe the methodology that we use to select which market instances will be tested for the criteria of Section III.

Our starting point is the 18 scenarios of the full factorial design described in the previous subsection.
For each of them, we compute screening indicators over the final evaluation window of each training run, after exploratory noise has decayed. The competitive benchmark is obtained by solving the same market-clearing problem~\eqref{market-clearing} under marginal-cost bidding, i.e. with all generators bidding their true marginal costs. Denoting the average realised price and total profit by $\bar{\lambda}^{obs}$ and $\Pi^{obs}$, and their competitive counterparts by $\bar{\lambda}^{comp}$ and $\Pi^{comp}$, we define
\begin{equation}
M^{\lambda} =
\frac{\bar{\lambda}^{obs}-\bar{\lambda}^{comp}}
{\bar{\lambda}^{comp}},
\end{equation}
\begin{equation}
M^{\Pi} =
\frac{\Pi^{obs}-\Pi^{comp}}
{\max\{\Pi^{comp},\epsilon_{\Pi}\}},
\end{equation}
where $\Pi=\sum_{i\in I}\pi_i$ denotes aggregate generator profit and $\epsilon_{\Pi}$ is a small denominator floor used to avoid degenerate ratios when competitive profits are close to zero.

To separate high-price outcomes from coordinated behaviour, the markup indicators are complemented by three dynamic indicators. First, price variance measures whether the learned market regime is stable enough for an intervention test to be meaningful. Second, action correlation and action-change correlation measure, respectively, whether agents maintain similar bidding levels or move their bidding strategies together over time. Third, Best Response Deviation Gain (BRDG) measures the short-run profitability of unilateral deviations. For agent $i$, it is defined as
\begin{equation}
\mathrm{BRDG}_i =
\frac{
\pi_i(c_i^{BR},\bar{c}_{-i})-\pi_i(\bar{c}_i,\bar{c}_{-i})
}{
\pi_i(\bar{c}_i,\bar{c}_{-i})
},
\end{equation}
where $\bar{c}_i$ denotes the bid prescribed by the learned policy of agent $i$, $\bar{c}_{-i}$ denotes the learned bids of all other agents, and $c_i^{BR}$ is the one-shot best response to $\bar{c}_{-i}$. A positive BRDG therefore indicates that the learned profile contains an unrealised short-run deviation incentive.

The selection favours cases that satisfy three practical requirements: (i) the learned regime is sufficiently stable to be meaningfully perturbed, (ii) the outcome is not a degenerate boundary solution in which all agents simply bid at the action-space limit, and (iii) there is a non-trivial unilateral deviation incentive that can be tested through the criteria of Section~\ref{sec:tacit_collusion}.
%
The selected cases are then evaluated using the Punishment of Deviation Criterion, the Unsustainability under Shortsight Criterion, and the Short-term Profitability of Deviations Criterion described in Section~\ref{sec:tacit_collusion}.

\begin{table}[t]
\caption{Scenarios selected for behavioural validation.}
\label{tab:selected_scenarios}
\centering
\scriptsize
\setlength{\tabcolsep}{2pt}
\begin{tabular*}{\columnwidth}{@{\extracolsep{\fill}}p{0.32\columnwidth} c c p{0.30\columnwidth}@{}}
\toprule
Scenario & Demand & Cost diff. & Rationale \\
\midrule
\texttt{Grid}--\texttt{LowDemand}--\texttt{HighCostDiff}
& Low
& High
& Structured non-stationary regime with non-trivial deviation incentives. \\

\texttt{Grid}--\texttt{LowDemand}--\texttt{MediumCostDiff}
& Low
& Medium
& Reference low-demand case with a more stationary learned profile. \\

\texttt{Grid}--\texttt{MediumDemand}--\texttt{HighCostDiff}
& Medium
& High
& Higher-demand case that remains non-boundary and strategically informative. \\
\bottomrule
\end{tabular*}
\end{table}

\subsection{MARL implementation} \label{marl-algorithm-section}

The action of agent $i$ at each instance $t$ is to select a bid vector $\bidvector = (c_{i,s}[t])_{s \in \segments}$, i.e. one declared marginal cost per segment $s$. This renders the action space of each agent to be of dimension $|\segments|$. To reduce this dimensionality and facilitate learning, we parameterize the bid vector through a \textit{supply function}.
Specifically, each agent $i$ maintains a linear supply function
\begin{equation}
    f(q_i; \beta_i) = \alpha_i + \beta_i \cdot \frac{q_i}{\sum_{s\in \segments}\segbound},
    \label{supply-function}
\end{equation}
which maps an output level $q_{i,s}$ to a declared marginal cost (bid) $\bid$. Note that this function is generally different than the agent's true cost function (defined in the previous subsection); it is an auxiliary bidding curve used internally to generate a consistent set of segment bids. The bid $\bid$ for segment $s$ is obtained by evaluating $f$ at 
the midpoint $\bar{q}_{i,s} = \sum_{s'\in \{0,1, ...s-1 \}} Q_{s'} + \frac{\segbound}{2}$ of the segment's capacity range:
\begin{equation}
    c_{i,s}[t] = f_i(\bar{q}_{i,s};\beta_i) = \alpha_i + \beta_i \cdot \frac{\bar{q}_{i,s}}{\sum_{s\in \segments}\segbound},
    \quad s \in \segments,
    \label{bid-parameterisation}
\end{equation}
where $\alpha_i$ is set to the agent's true marginal cost and $\beta_i$ is the parameter to be learned by the agent.
This defines a deterministic mapping $\beta_i[t] \mapsto \bidvector$, so that the full bid 
vector is determined by a single scalar parameter. The agent's internal action at each instance 
$t$ is therefore the choice of $\beta_i[t] \in \mathbb{R}_{\geq 0}$, which is then translated 
into the game action $\bidvector$ via \eqref{bid-parameterisation}. 

\begin{remark}
    This parameterization reduces the dimensionality of an agent's action space from $|\segments|$ to $1$. This is a choice that only makes the validation of our \textit{Hypothesis} more difficult: by reducing the agents' degrees of freedom, we reduce the richness of their policies, possibly excluding instances of collusive joint policies. Thus, if collusive policies are not discovered with this model, it does not necessarily mean that they don't exist. But if collusive policies \textit{are} discovered with this model, then the validation of the \textit{Hypothesis} holds also for the general model where each agent's action comprises an arbitrary choice for each segment's bid. \hfill $\square$
\end{remark}

In the experiments, the supply function of each generator is discretised into 12 equal-capacity bid segments. The bidding parameter $\beta_i$ is constrained to the interval $[0,4000]$, where the upper bound corresponds to the maximum admissible value of the slope parameter. The upper bound is consistent with the market price cap of 4000~EUR/MWh.

The agents are trained using a multi-agent implementation of TD3 under the centralized-training/decentralized-execution paradigm. Each generator is represented by a deterministic actor that maps its private history to the bidding parameter used to construct its offer. During training, critics are allowed to condition on joint market information in order to stabilize learning in the non-stationary multi-agent environment. During execution and evaluation, however, only the decentralized actors are used. The agents therefore do not communicate when submitting bids.

Each of the 18 scenarios is trained for 100,000 market instances, organised as 50 episodes of 2,000 instances. The first 4,000 instances are used as a warm-up phase, during which agents execute randomly generated actions to populate the replay buffer. Exploration is introduced through action noise and is gradually removed over the course of training, so that the final evaluation window reflects the learned deterministic policies rather than exploratory behaviour. All screening indicators reported in the previous subsection are computed over the final 10,000 market instances.
The main hyperparameters are reported in Table~\ref{tab:marl_hyperparameters}. 

\begin{table}[t]
\caption{Main MARL hyperparameters.}
\label{tab:marl_hyperparameters}
\centering
\scriptsize
\setlength{\tabcolsep}{2pt}
\begin{tabular*}{\columnwidth}{@{\extracolsep{\fill}}l c@{}}
\toprule
Parameter & Value \\
\midrule
Algorithm & Multi-agent TD3 \\
Bid segments & 12 \\
Bidding parameter bounds & $[0,4000]$ \\
Training length & 100,000 market instances \\
Episodes / horizon & $50 \times 2,000$ \\
Evaluation window & Final 10,000 instances \\
Warm-up phase & 4,000 instances \\
Discount factor & 0.99 \\
Replay buffer size & 50,000 \\
Batch size & 256 \\
Actor / critic learning rate & $10^{-4}$ / $3\cdot 10^{-4}$ \\
Hidden layers & $[256,256]$ \\
Exploration decay & 90,000 instances \\
\bottomrule
\end{tabular*}
\end{table}

\section{Results} \label{sec:results}

\subsection{Screening Outcomes}

\begin{figure*}[!t]
    \centering

    \subfloat[\label{fig:markup_no_grid}]{
        \includegraphics[width=0.40\textwidth]
        {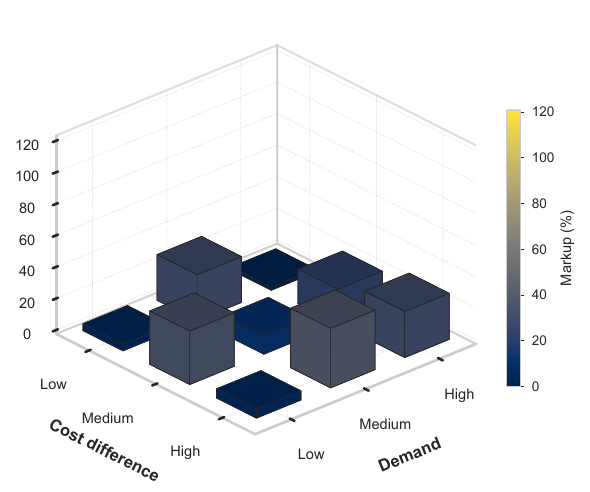}
    }
    \hfill
    \subfloat[\label{fig:markup_grid}]{
        \includegraphics[width=0.40\textwidth]
        {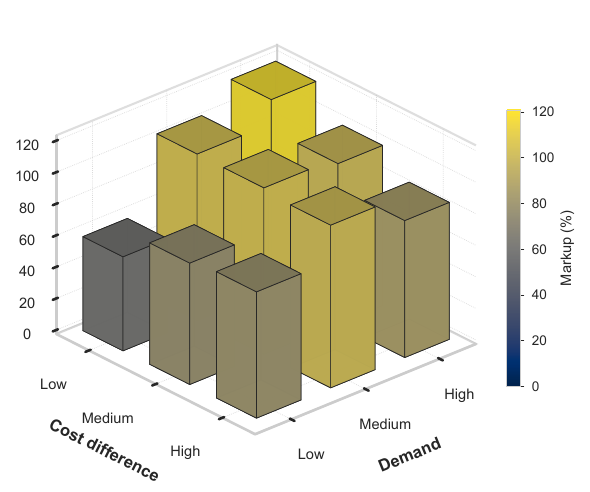}
    }

    \caption{Price markups across the nine demand and marginal-cost heterogeneity combinations: (a) \texttt{NoGrid} scenarios without active transmission constraints and (b) \texttt{Grid} scenarios with active transmission constraints.}
    \label{fig:markup_comparison}
\end{figure*}

Among the screening indicators introduced in Section~\ref{sec:experimental_setting}, Fig.~\ref{fig:markup_comparison} reports the price markup across the full set of 18 market environments. Markups are generally modest in the \texttt{NoGrid} cases and do not exhibit a clear ordering across demand levels or cost heterogeneity. In contrast, the \texttt{Grid} cases produce substantially higher supra-competitive markups and display a systematic pattern in which higher demand and lower marginal-cost heterogeneity are associated with higher markups. This indicates that transmission constraints materially amplify market power and make the resulting market outcomes more sensitive to the underlying demand and cost structure. At the same time, elevated markups alone cannot distinguish coordinated behaviour from market power arising directly from network constraints.

Applying the screening procedure described in Section~\ref{sec:experimental_setting} leaves three non-boundary \texttt{Grid} cases for behavioural validation: \texttt{Grid}-\texttt{LowDemand}-\texttt{HighCostDiff}, \texttt{Grid}-\texttt{LowDemand}-\texttt{MediumCostDiff}, and \texttt{Grid}-\texttt{MediumDemand}-\texttt{HighCostDiff}. These cases all exhibit supra-competitive outcomes while differing in demand conditions, cost asymmetry, and learned strategic dynamics.

\subsection{Punishment of Deviations}

\begin{figure*}[t]
    \centering
    \includegraphics[width=0.95\textwidth]{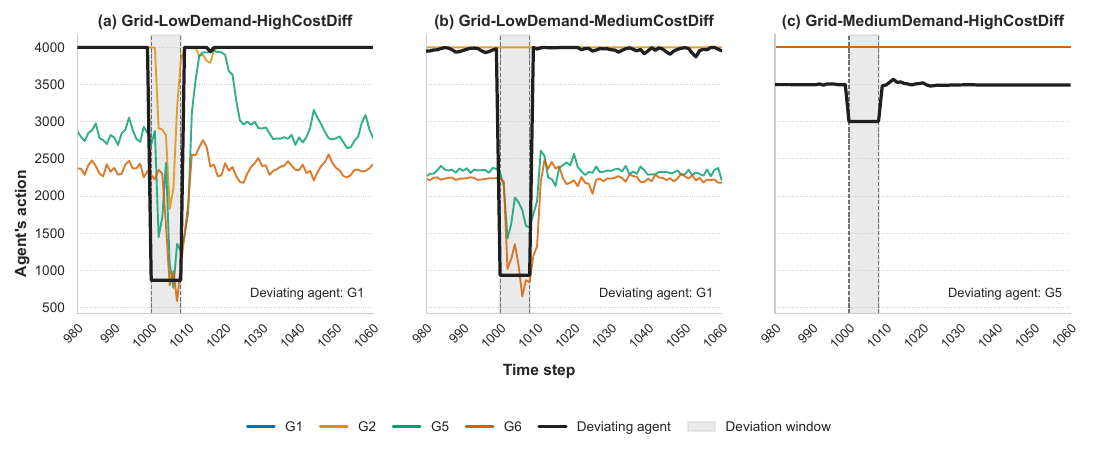}
    \caption{Agents' bidding actions during the Punishment of Deviation test for the three selected scenarios. The shaded region denotes the forced-deviation window, and the black line identifies the deviating agent.}
    \label{fig:punish_strategic_action}
\end{figure*}

Figure~\ref{fig:punish_strategic_action} shows the bidding responses to the forced deviation in the three selected scenarios. In \texttt{Grid}-\texttt{LowDemand}-\texttt{HighCostDiff}, the deviation of $G_1$ triggers a pronounced response from the remaining agents, which temporarily lower their bidding slopes and subsequently return towards their pre-deviation strategies, producing a clear punish-and-forgive pattern. A similar pattern emerges in \texttt{Grid}-\texttt{LowDemand}-\texttt{MediumCostDiff}, although the response is asymmetric: $G_5$ and $G_6$ react strongly, while $G_2$ remains close to its pre-deviation strategy. A plausible explanation for this difference is the change in the residual-supply structure induced by the different relative competitiveness of $G_5$ and $G_6$, rather than any change in $G_2$'s own marginal cost, which is identical in the two scenarios. Under \texttt{HighCostDiff}, the very competitive $G_6$ is strongly constrained by its network position while $G_5$ is a relatively costly substitute, increasing the marginal value of $G_2$'s available capacity and making its participation in the punishment more important. Under \texttt{MediumCostDiff}, $G_5$ becomes more competitive and $G_6$ retains more marginal supply headroom, so their combined response can substitute more effectively for $G_2$, reducing the marginal value of its participation and creating a natural free-riding incentive. In \texttt{Grid}-\texttt{MediumDemand}-\texttt{HighCostDiff}, by contrast, the deviation of $G_5$ does not induce a comparable coordinated reduction in competitors' bidding slopes, and no clear punish-and-forgive pattern is observed.

\begin{figure*}[t]
    \centering
    \includegraphics[width=0.95\textwidth]{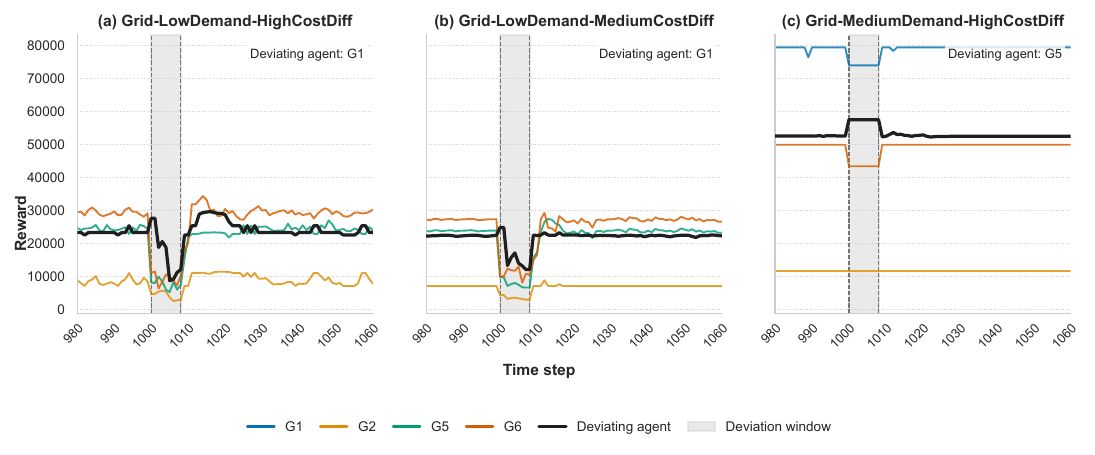}
    \caption{Agents' rewards during the Punishment of Deviation test for the three selected scenarios. The shaded region denotes the forced-deviation window, and the black line identifies the deviating agent.}
    \label{fig:punish_reward}
\end{figure*}

The corresponding reward trajectories in Fig.~\ref{fig:punish_reward} show whether these strategic responses impose an economic penalty on the deviating agent. In \texttt{Grid}-\texttt{LowDemand}-\texttt{HighCostDiff}, the response of the other agents reduces the profitability of $G_1$'s deviation, with rewards subsequently recovering as the agents return towards the previous regime. In \texttt{Grid}-\texttt{LowDemand}-\texttt{MediumCostDiff}, the response of $G_5$ and $G_6$ is likewise sufficient to penalize the deviator despite the weak reaction of $G_2$, supporting the interpretation that $G_2$ is not pivotal to punishment in this case. In \texttt{Grid}-\texttt{MediumDemand}-\texttt{HighCostDiff}, the absence of a comparable strategic response is reflected in the deviator retaining the benefit of its deviation, providing no evidence of an effective punishment mechanism. The Punishment of Deviation criterion is therefore supported in both \texttt{LowDemand} scenarios, although with an asymmetric response in the \texttt{MediumCostDiff} case, and is not supported in \texttt{Grid}-\texttt{MediumDemand}-\texttt{HighCostDiff}.

\subsection{Shortsight Ablation}

\begin{figure*}[t]
    \centering
    \includegraphics[width=0.95\textwidth]{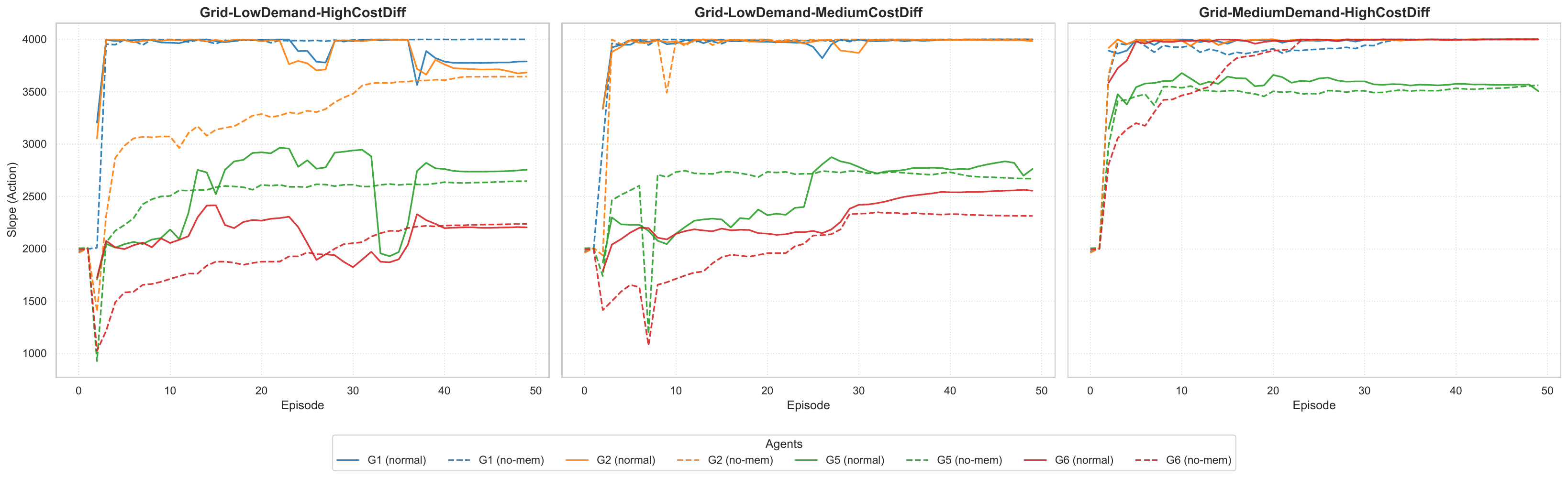}
    \caption{Evolution of episode-average bidding slopes under the baseline MARL configuration and the shortsight ablation for the three selected scenarios. Solid lines denote the baseline agents and dashed lines the ablated agents.}
    \label{fig:shortsight_training}
\end{figure*}

Figure~\ref{fig:shortsight_training} compares the evolution of episode-average bidding slopes under the baseline MARL configuration and the shortsight ablation. Contrary to a simple collapse-to-competition interpretation, removing memory and long-horizon incentives does not systematically reduce the learned bidding slopes: the magnitude and direction of the changes vary across agents and scenarios, while the \texttt{Grid}-\texttt{MediumDemand}-\texttt{HighCostDiff} case remains particularly close to the baseline. A plausible explanation is that a substantial part of the on-path supra-competitive outcome is supported by static, network-induced market power. Transmission constraints, local scarcity, and generator pivotality remain unchanged after the ablation, so high bidding slopes may remain profitable even for nearly myopic agents. The stationarity of demand, costs, and network parameters further allows the learned policies to stabilize around similar actions without relying on long histories of past market outcomes. The ablated runs also converge more smoothly, consistent with weaker inter-agent feedback loops. Hence, the similarity of the on-path outcomes does not imply that the underlying strategic mechanism is unchanged; the relevant distinction is between the level of the learned regime and the agents' ability to condition their behaviour on deviations from it.

\begin{figure*}[t]
    \centering
    \includegraphics[width=0.95\textwidth]{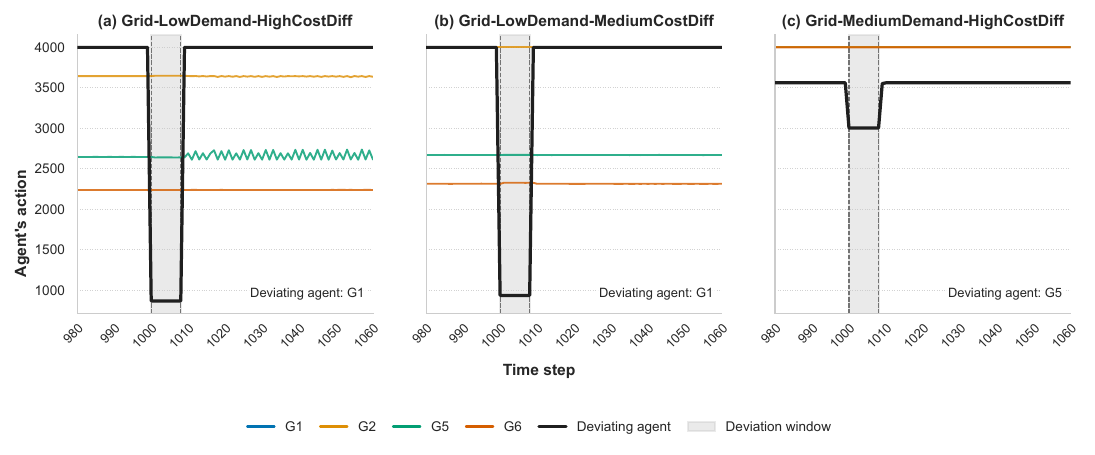}
    \caption{Agents' bidding actions during the Punishment of Deviation test under the shortsight ablation for the three selected scenarios. The shaded region denotes the forced-deviation window, and the black line identifies the deviating agent.}
    \label{fig:punish_forgive_action_no_memory}
\end{figure*}

This distinction becomes apparent when the forced-deviation experiment is repeated under the shortsight ablation, as shown in Fig.~\ref{fig:punish_forgive_action_no_memory}. In both \texttt{LowDemand} scenarios, the pronounced punishment-and-forgiveness responses observed under the baseline disappear, with competitors' bidding slopes remaining largely unchanged following the forced deviation. Removing history eliminates the informational basis for conditioning current actions on a previous deviation, while the near-myopic discount factor removes much of the incentive to incur a short-run punishment cost in order to restore a more profitable future regime. In \texttt{Grid}-\texttt{MediumDemand}-\texttt{HighCostDiff}, where no clear punishment response was present under the baseline, the ablation likewise produces no evidence of an enforcement mechanism. The shortsight ablation therefore affects primarily the off-path enforcement mechanism rather than necessarily the on-path level of bids. It consequently provides supporting evidence for a repeated-game component in the two \texttt{LowDemand} scenarios through the disappearance of punishment, while the persistence of similar on-path bidding levels indicates that their supra-competitive outcomes are also partly supported by static market power; no comparable evidence is obtained for \texttt{Grid}-\texttt{MediumDemand}-\texttt{HighCostDiff}.

\subsection{Iterative Best Responses}

\begin{figure*}[t]
    \centering
    \includegraphics[width=\textwidth]{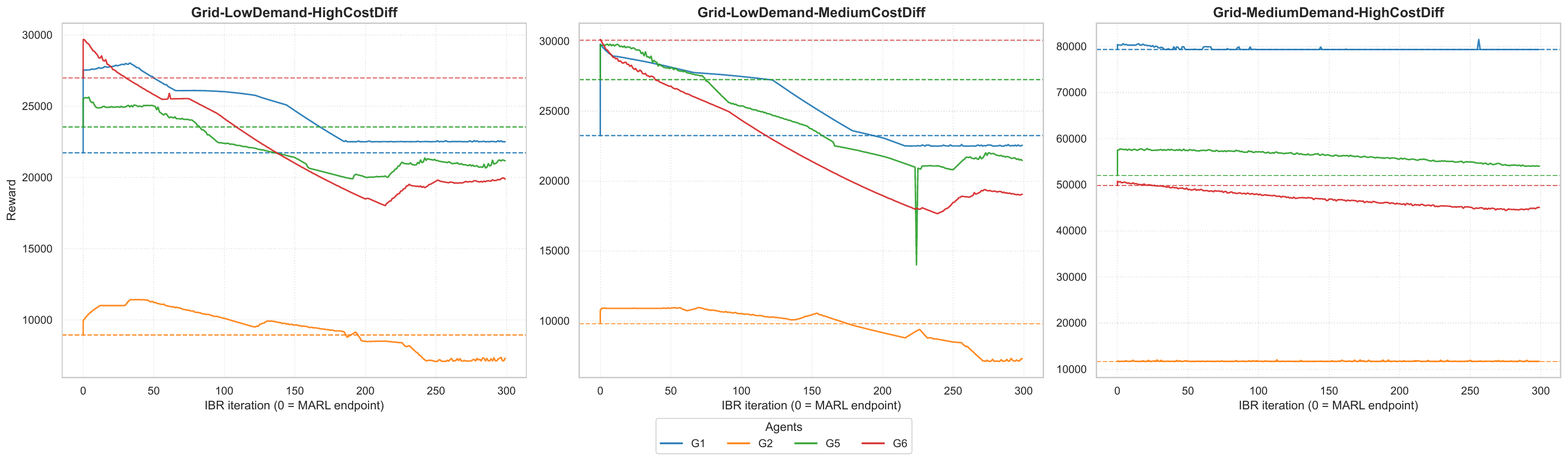}
    \caption{Agent reward trajectories under Iterative Best Response (IBR), initialized from the learned MARL outcomes. Iteration 0 corresponds to the MARL endpoint, while the horizontal dashed lines indicate the corresponding MARL reward levels.}
    \label{fig:ibr_trajectories}
\end{figure*}

Figure~\ref{fig:ibr_trajectories} shows the reward trajectories obtained when the learned MARL outcomes are used as the starting point for Iterative Best Response. In \texttt{Grid}-\texttt{LowDemand}-\texttt{HighCostDiff}, unilateral best responses provide short-run profit opportunities, but the subsequent IBR dynamics move the agents towards reward levels that are generally below those sustained under MARL. The pattern is even clearer in \texttt{Grid}-\texttt{LowDemand}-\texttt{MediumCostDiff}, where the initially profitable deviations are followed by lower reward levels for all agents relative to their MARL benchmarks. These two cases are therefore consistent with a regime in which unilateral deviation is attractive in the short run, while the competitive adaptation induced by continued best responses is less profitable. In \texttt{Grid}-\texttt{MediumDemand}-\texttt{HighCostDiff}, by contrast, the IBR trajectories do not exhibit the same combination of profitable short-run deviations and a systematically less profitable subsequent regime. The Short-term Profitability of Deviations criterion is therefore supported in the two \texttt{LowDemand} scenarios and not supported in \texttt{Grid}-\texttt{MediumDemand}-\texttt{HighCostDiff}.

\subsection{Synthesis of Behavioural Evidence}

\begin{table}[t]
\caption{Summary of behavioural evidence for Tacit Collusion.}
\label{tab:collusion_synthesis}
\centering
\scriptsize
\setlength{\tabcolsep}{1pt}
\begin{tabular*}{\columnwidth}{@{\extracolsep{\fill}}p{0.29\columnwidth} c c c p{0.23\columnwidth}@{}}
\toprule
Scenario & Punish. & Shortsight & IBR & Interpretation \\
\midrule
\texttt{Grid}--\texttt{LowDemand}--\texttt{HighCostDiff}
& Supportive
& Supportive
& Supportive
& Collusion-consistent \\

\texttt{Grid}--\texttt{LowDemand}--\texttt{MediumCostDiff}
& Supportive
& Supportive
& Supportive
& Collusion-consistent \\

\texttt{Grid}--\texttt{MediumDemand}--\texttt{HighCostDiff}
& Not supp.
& Not supp.
& Not supp.
& Structural market power \\
\bottomrule
\end{tabular*}
\end{table}

Table~\ref{tab:collusion_synthesis} summarizes the evidence obtained from the three behavioural criteria.
Taken together, the three behavioural tests provide mutually consistent evidence that the two \texttt{LowDemand} learned regimes contain a repeated-game component consistent with Tacit Collusion. The persistence of supra-competitive on-path bidding after the shortsight ablation nevertheless indicates that these outcomes are not driven by repeated-game incentives alone, but are also supported by structural, network-induced market power. \texttt{Grid}-\texttt{MediumDemand}-\texttt{HighCostDiff} provides the contrasting case: despite exhibiting a supra-competitive market outcome, it shows neither a punishment response, an ablation-sensitive enforcement mechanism, nor an IBR pattern consistent with the collusion criteria. Its elevated markup is therefore more plausibly attributed to structural market power than to Tacit Collusion. Overall, the results show that similar supra-competitive outcomes can arise from qualitatively different strategic mechanisms, which cannot be distinguished from price levels alone.

\section{Conclusions} \label{sec:conclusion}

This paper investigated whether Tacit Collusion can emerge in electricity markets whose participants delegate their bidding strategies to autonomous, learning-based algorithms. 
Strategic bidding was modeled as a repeated game with imperfect public monitoring, with agents' learning-based policies modeled via multi-agent reinforcement learning.
Different cases of market characteristics were filtered based on susceptibility for Tacit Collusion and the most suspicious ones were tested against three Tacit Collusion criteria. Our experimental results show that the danger of Tacit Collusion in electricity markets is realistic: particularly under binding network constraints, agents learn to sustain supra-competitive outcomes that satisfy several independent indicators of Tacit Collusion, despite never being instructed or designed to collude.
Our results should not be taken as conclusive but rather as an alarm bell that encourages extensive research on this topic. Specifically, investigating the learning mechanics that make Tacit Collusion emergent and investigating market-design countermeasures constitute important future work directions.

\bibliographystyle{IEEEtran}
\bibliography{references}

\end{document}